\documentclass{article}

\usepackage{arxiv}

\usepackage[utf8]{inputenc} 
\usepackage[T1]{fontenc}    
\usepackage{hyperref}       
\usepackage{url}            
\usepackage{booktabs}       
\usepackage{amsfonts}       
\usepackage{nicefrac}       
\usepackage{microtype}      
\usepackage{lipsum}		
\usepackage{graphicx}
\usepackage{natbib}
\usepackage{doi}

\usepackage{color}
\usepackage{multirow}
\usepackage{booktabs}
\usepackage{tabularray}
\usepackage{xspace}
\usepackage{colortbl}
\usepackage{hhline}
\usepackage{arydshln}
\usepackage{tcolorbox}
\usepackage{subcaption}

\newcommand{\codem}{\textsc{CodeM}\xspace}

\title{Can Programming Languages Boost Each Other\\ via Instruction Tuning?}

\author{
Daoguang Zan\textsuperscript{$\mathcal{y}$}\footnotemark[1]~~Ailun Yu\textsuperscript{$\mathcal{x}$}\thanks{The first two authors contributed equally to this work.}~~Bo Shen\textsuperscript{$\mathcal{z}$}~~Jiaxin Zhang\textsuperscript{$\mathcal{z}$}~~Taihong Chen\textsuperscript{$\mathcal{z}$}~~Bing Geng\textsuperscript{$\mathcal{z}$}~~Bei Chen\textsuperscript{\P}\\
~~\textbf{Jichuan Ji\textsuperscript{$\mathcal{z}$}}~~\textbf{Yafen Yao\textsuperscript{$\mathcal{z}$}}~~\textbf{Yongji Wang\textsuperscript{$\mathcal{y}$}}~~\textbf{Qianxiang Wang\textsuperscript{$\mathcal{z}$}}
\\
\textsuperscript{$\mathcal{y}$}Institute of Software, Chinese Academy of Science \\
\textsuperscript{$\mathcal{x}$}Peking University \\
\textsuperscript{$\mathcal{z}$}Huawei Co., Ltd. \\
\textsuperscript{\P}Independent Researcher \\
\texttt{daoguang@iscas.ac.cn; yuailun@pku.edu.cn}
}

\date{}

\renewcommand{\headeright}{Technical Report}
\renewcommand{\undertitle}{Technical Report}
\renewcommand{\shorttitle}{}

\hypersetup{
pdftitle={\codem: Can Programming Languages Boost Each Other via Instruction Tuning?},
pdfsubject={q-bio.NC, q-bio.QM},
pdfauthor={Daoguang Zan},
pdfkeywords={Code Large Language Model, Instruction Tuning, More},
}

\begin{document}
\maketitle

\begin{abstract}

When human programmers have mastered a programming language, it would be easier when they learn a new programming language.
In this report, we focus on exploring whether programming languages can boost each other during the instruction fine-tuning phase of code large language models. 
We conduct extensive experiments of $8$ popular programming languages (Python, JavaScript, TypeScript, C, C++, Java, Go, HTML) on StarCoder.
Results demonstrate that programming languages can significantly improve each other.
For example, \codem-Python $15$B trained on Python is able to increase Java by an absolute $17.95$\% pass$@1$ on HumanEval-X.
More surprisingly, we found that \codem-HTML $7$B trained on the HTML corpus can improve Java by an absolute $15.24$\% pass$@1$.
Our training data is released at \url{https://github.com/NL2Code/CodeM}.

\end{abstract}


\section{Introduction}

Code large language models (code LLMs) are blooming recently~\citep{nl2code}.
A lot of code LLMs are released in succession, e.g., Codex~\citep{codex}, AlphaCode~\citep{alphacode}, PaLM-Coder~\citep{palm}, CodeGen~\citep{codegen}, CodeGeeX~\citep{codegeex}, StarCoder~\citep{starcoder}, and Code Llama~\citep{codellama}.
Owing to their amazing code generation performance, code LLMs have attracted considerable attention from both academic and industrial circles.
Recent works~\citep{instructgpt} have witnessed the instruction tuning technique that can teach LLMs how to follow instructions.
In the realm of code generation, WizardCoder~\citep{wizardcoder} and PanGu-Coder2~\citep{pangucoder2} also adopt this technique to elicit their code generation capabilities.
Although some code LLMs, such as CodeGen-Multi~\cite{codegen} and StarCoder-base~\cite{starcoder}, are trained on corpora spanning multiple programming languages, the interplay among these languages remains unexplored. 
In programming practice, once a human programmer has mastered a programming language, it would be easier to learn a new one due to the homogeneity between programming languages.
Motivated by this, we would like to explore whether different programming languages can boost each other during instruction fine-tuning of code LLMs.

To explore this idea, we craft the training corpus for each of $8$ popular programming languages (Python, JavaScript, TypeScript, C, C++, Java, Go, HTML), where each language includes about $9$K programming exercises.
We train StarCoder $7$B using the instruction tuning technique on each programming language corpus separately, and test the performance of each fine-tuned model across every programming language.
Our findings reveal that programming languages can significantly boost each other.
Meanwhile, we found that the improvement margin of different programming languages to each other is related to the language similarity between them.
For example, \codem-JavaScript 7B trained on JavaScript data can yield an absolute $11.80$\% pass$@1$ improvement in TypeScript.
More interestingly, \codem-HTML 7B trained on the markup language HTML also can achieve an absolute $15.24$\% pass$@1$ improvement in Java.

In a nutshell, our contributions can be listed as follows:
(1) Our findings suggest that programming languages can significantly boost each other during code LLMs' instruction fine-tuning phase.
(2) We glean valuable insights on the correlation between multiple programming languages, paving the way for future research on code generation.
(3) We will make our training data publicly available.

\section{Methodology}
\subsection{Crafting Training Corpus of Eight Programming Languages} \label{sec:craft_multiligual_data}
We select $8$ popular programming languages and construct their training data separately.
Our selected languages include Python, JavaScript, TypeScript, C, C++, Java, Go, and HTML, covering diverse types such as procedure-oriented, object-oriented, script, and even markup languages.
For each programming language, we construct its training data containing about $9$K data pairs.
Each pair includes both an instruction describing the programming problem and its corresponding response.
One practical example of HTML is shown in Figure~\ref{fig:training_example}.

Based on these selected languages, we construct a series of monolingual datasets.
We start from the dataset of CodeAlpaca 20K\footnote{\url{https://huggingface.co/datasets/sahil2801/CodeAlpaca-20k}}, and extract those Python-related data to form our seed instruction set.
Then for each selected programming language, we evolve existent instructions in the seed instruction set to get corresponding new ones by prompting OpenAI's GPT-3.5\footnote{\url{https://platform.openai.com/docs/models/gpt-3-5}}.
For all the selected languages except HTML, we adopt an in-depth evolution~\citep{wizardlm}, by asking GPT-3.5 to rewrite the seed instruction (Python) into a more complicated version relevant to the target language (Python, JavaScript, TypeScript, C, C++, Java, or Go).
However, for HTML, we adopt in-breadth evolution to produce a brand-new HTML-related instruction, since HTML (markup language) is too different from other languages (non-markup languages).

\begin{figure}
    \centering
\begin{tcolorbox}[left=2pt,right=2pt,top=0pt,bottom=0pt]
\textcolor{gray}{\# Instruction}
\\
Design a web page that displays a message when loaded. The message should be "Hello, World". Use HTML code to achieve this.
\\
\\
\textcolor{gray}{\# Response}
\\
<!DOCTYPE html>\\
<html>\\
<head>\\
\hspace*{2em}<title>Hello World</title>\\
</head>\\
<body>\\
\hspace*{2em}<h1>Hello, World</h1>\\
</body>\\
</html>
\end{tcolorbox}
\caption{A HTML training example of our crafted instruction-answer pairs.}
\label{fig:training_example}
\end{figure}

\subsection{Instruction Tuning}
Code pre-trained models such as Codex~\citep{codex} and StarCoder~\citep{starcoder} store a wealth of code knowledge.
However, these models only support left-to-right code generation based on context, as they are trained solely on plain code snippets.
Of late, the instruction tuning techniques~\citep{instructgpt,wizardcoder,pangucoder2} are proposed, which can enhance the model's capabilities of following instructions so as to enable chat features.
During instruction tuning, we train StarCoder using the prompt in Figure~\ref{fig:prompt_training} to obtain our \codem.
We use DeepSpeed to accelerate the training of \codem with \texttt{fp16} enabled.
Additionally, we set the batch size to $2$ per GPU, the learning rate to $2$e-$5$ with a cosine annealing schedule, the gradient accumulation steps to $4$, and the warmup steps to $30$.
After instruction tuning, we use the prompt in Figure~\ref{fig:prompt_inferencing} to do the inference on downstream tasks across various programming languages.
For inference, we adopt the greedy decoding strategy for sampling.
Given that \codem is a chat-style model, the responses it generates often contain elements beyond just codes, which typically makes them non-executable.
So, we extract the code snippets from the generated response to evaluate the performance of code generation.

\begin{figure}
    \centering
\begin{tcolorbox}[left=2pt,right=2pt,top=0pt,bottom=0pt]
Below is an instruction that describes a task, paired with an input that provides further context. Write a response that appropriately completes the request.
\\
\\
\#\#\# Instruction:
\\
\{problem\}
\\
\\
\#\#\# Response:
\\
\{response\}
\end{tcolorbox}
\caption{Prompt format of instruction tuning. \texttt{\{problem\}} and \texttt{\{response\}} refer to the instruction and answer obtained in Section~\ref{sec:craft_multiligual_data}.}
\label{fig:prompt_training}
\end{figure}

\begin{figure}
    \centering
\begin{tcolorbox}[left=2pt,right=2pt,top=0pt,bottom=0pt]
Below is an instruction that describes a task. Write a response that appropriately completes the request.
\\
\\
\#\#\# Instruction:
\\
Finish the \{language\} code for this problem:
\\
\{problem\}
\\
\\
\#\#\# Response:
\\
\{signature\}
\end{tcolorbox}
\caption{Prompt format of inference. \texttt{\{language\}}, \texttt{\{problem\}}, and \texttt{\{signature\}} represent the downstream programming language, the given programming problem, and the function header, respectively.}
\label{fig:prompt_inferencing}
\end{figure}

\section{Experiments}

\subsection{Evaluation Setup}
\subsubsection{Benchmarks and Baselines}
We use HumanEval-X~\citep{codegeex} to evaluate the multilingual abilities of models in Python, JavaScript, C++, Java, and Go.
HumanEval-X is crafted by adapting HumanEval~\citep{codex} (Python) to other programming languages.
Following the same approach as HumanEval-X, we also create two new versions of HumanEval: HumanEval-C and HumanEval-TypeScript.
Note that HumanEval can not directly be adapted to markup languages such as HTML, so our downstream evaluation languages do not include HTML.

The primary baseline for all language versions of \codem is their base model StarCoder.
We analyze whether \codem trained on language A can improve language B, in which case the baselines are \codem directly trained on language B.

\subsubsection{Metrics}
We adopt pass$@1$ as our metric to evaluate all the models.
Each model generates one answer using the greedy decoding strategy for each programming task, and the answer would be executed upon the given test cases.
Only when all the test cases are passed, the programming task can be considered solved with the generated code.
In this setting, pass$@1$ can be formulated as $\frac{|P_{c}|}{|P|}$, where $|P|$ denotes the total number of programming tasks in HumanEval and $|P_{c}|$ represents the number of solved tasks. 
In essence, the pass$@1$ metric we use can be considered as the accuracy.

\subsection{Results}

\subsubsection{Main Results}

Table~\ref{tab:main_results} shows the performance of \codem, which are a series of models trained on monolingual datasets of eight languages respectively, across different language versions of HumanEval.
As we can see, all \codem models outperform their base model StarCoder $7$B across all programming languages by a large margin.
Also, we found that programming languages can boost each other significantly.
For example, \codem-Python trained solely on Python corpus is able to improve HumanEval-Java by an absolute $14.03$\% pass$@1$.
This finding reveals the inherent commonalities among different programming languages.
More surprisingly, \codem-HTML boosts HumanEval-Java by an absolute $15.24$\% pass$@1$, even exceeding \codem-Java.
Similarly, \codem-C++ beats \codem-C on HumanEval-C, and \codem-JavaScript beats \codem-TypeScript on HumanEval-Typescript.
Drawing upon these observations, we conjecture that the improvement in multilingual code generation performance is predominantly due to instruction tuning unlocking the model's inherent potential, such as natural or programming language understanding and following-instruction capabilities, rather than merely incorporating new knowledge.
In addition to training \codem on a monolingual training corpus, we further construct a $9$K multilingual training set covering $8$ programming languages.
Although each language comprises only a small amount (\textasciitilde $1.2$K) of training instances, experimental findings suggest that \codem-Mixed excels in all languages, even surpassing \codem-Python on HumanEval-Python and \codem-Java on HumanEval-Java.
This suggests that it is possible to yield superior code generation performance by leveraging multilingual data in instruction tuning, without harming the generalization of the model.

We also conduct experiments on StarCoder $15$B to verify the effectiveness of \codem.
Specifically, we obtain $108$K Python training data following WizardCoder~\citep{wizardcoder}, and finetune StarCoder $15$B to get \codem-Python.
The results are shown in Table~\ref{tab:results_15B}.
\codem-Python achieves state-of-the-art performance on HumanEval-Python with $64.63$\% pass$@1$, compared with other models of the same scale.
\codem-Python also gets a tremendous improvement in the generation of other programming languages.
For instance, it improves Java and JavaScript by an absolute $17.95$\% and $16.77$\% pass$@1$, respectively.

\subsubsection{Closer Analysis}

\begin{figure}[t]
	\centering
	\begin{subfigure}[t]{0.45\linewidth}
		\centering
		\includegraphics[width=\linewidth]{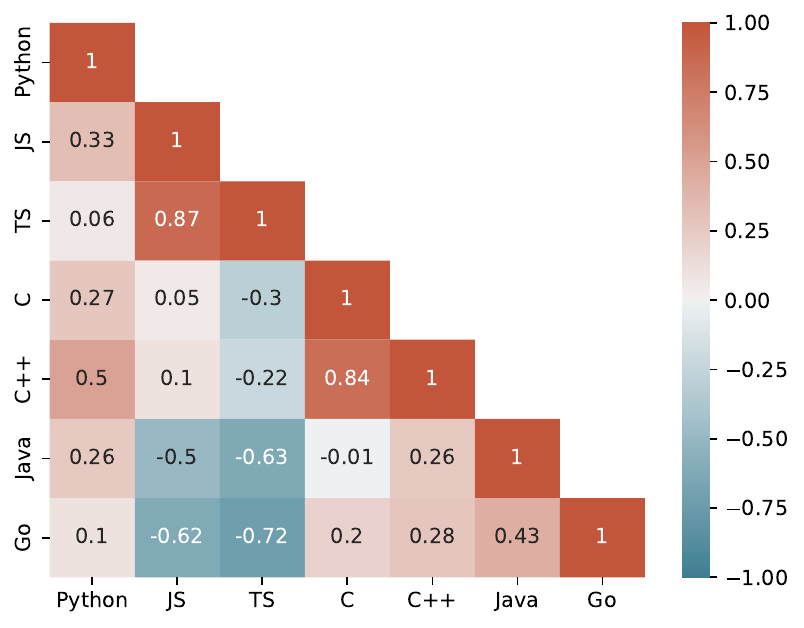}
            \caption{}
		\label{subfig:correlation_col}
	\end{subfigure}
        \hspace{0.5cm}
	\begin{subfigure}[t]{0.45\linewidth}
		\centering
		\includegraphics[width=\linewidth]{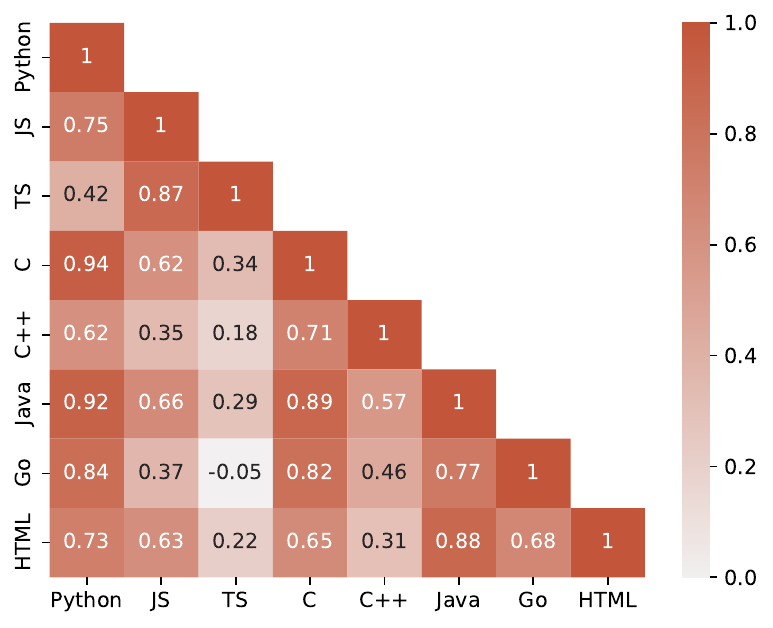}
            \caption{}
		\label{subfig:correlation_row}
	\end{subfigure}
        \caption{Correlations between different programming languages. We regard the data in Table~\ref{tab:main_results} as a matrix, and use ``\texttt{df.corr()}'' from the Pandas library to compute the correlation between different programming languages. The correlation results before and after ``\texttt{df.T}'' are presented in (a) and (b), respectively.}
	\label{fig:correlation}
\end{figure}

We analyze the correlation between different programming languages.
As illustrated in Figure~\ref{fig:correlation} (a), the improvement of code generation performance is sensitive to training corpus of different programming languages.
Moreover, we found that C and C++ can boost each other more significantly, which is the same for JavaScript and TypeScript.
It is reasonable because these languages are correlated to each other in language design, sharing some common syntax and grammar.
Figure~\ref{fig:correlation} (b) shows that training on each programming language can boost the code generation performance of all other languages.
We can see that the correlation values in Figure~\ref{fig:correlation} (b) are mostly all positive, implying that the improvement trend of different language brought by one monolingual training corpus is relatively similar.

\begin{table}
\centering
\caption{Pass$@1$ (Accuracy) of StarCoder $7$B and \codem trained on various programming languages. The numbers in \textcolor{red}{red} represent the absolute increase compared to StarCoder 7B.}
\label{tab:main_results}
\begin{tabular}{l|lllllll} 
\toprule
\multirow{2}{*}{\textbf{Model}}  & \multicolumn{7}{c}{\textbf{HumanEval-Multilingual}}                                                                                                                                                                                                                                                                                                                                                         \\
                                 & Python                                                 & JavaScript                                             & TypeScript                                             & C                                                     & C++                                                    & Java                                                   & Go                                                     \\ 
\hline\hline
StarCoder 7B                     & 26.83                                                  & 24.39                                                  & 28.57                                                  & 24.69                                                 & 25.61                                                  & 23.17                                                  & 24.39                                                  \\ 
\hdashline
\codem-Python     & 38.41\textsuperscript{\textcolor{red}{11.58}}          & 34.76\textsuperscript{\textcolor{red}{10.37}}          & 33.54\textsuperscript{\textcolor{red}{4.97}}           & 29.01\textsuperscript{\textcolor{red}{4.32}}          & 34.15\textsuperscript{\textcolor{red}{8.54}}           & 37.20\textsuperscript{\textcolor{red}{14.03}}          & 27.44\textsuperscript{\textcolor{red}{3.05}}           \\
\codem-JavaScript & 37.20\textsuperscript{\textcolor{red}{10.37}}          & \textbf{40.24}\textsuperscript{\textcolor{red}{15.85}} & \textbf{40.37}\textsuperscript{\textcolor{red}{11.80}} & 27.78\textsuperscript{\textcolor{red}{3.09}}          & 32.93\textsuperscript{\textcolor{red}{7.32}}           & 34.76\textsuperscript{\textcolor{red}{11.59}}          & 26.22\textsuperscript{\textcolor{red}{1.83}}           \\
\codem-TypeScript & 33.54\textsuperscript{\textcolor{red}{6.71}}           & 37.80\textsuperscript{\textcolor{red}{13.41}}          & 37.27\textsuperscript{\textcolor{red}{8.70}}           & 30.25\textsuperscript{\textcolor{red}{5.56}}          & 30.49\textsuperscript{\textcolor{red}{4.88}}           & 28.05\textsuperscript{\textcolor{red}{4.88}}           & 25.61\textsuperscript{\textcolor{red}{1.22}}           \\
\codem-C          & 39.63\textsuperscript{\textcolor{red}{12.8}}           & 37.20\textsuperscript{\textcolor{red}{12.81}}          & 32.30\textsuperscript{\textcolor{red}{3.73}}           & 32.10\textsuperscript{\textcolor{red}{7.41}}          & 35.37\textsuperscript{\textcolor{red}{9.76}}           & 38.41\textsuperscript{\textcolor{red}{15.24}}          & 28.66\textsuperscript{\textcolor{red}{4.27}}           \\
\codem-C++        & 34.57\textsuperscript{\textcolor{red}{7.74}}           & 35.37\textsuperscript{\textcolor{red}{10.98}}          & 32.30\textsuperscript{\textcolor{red}{3.73}}           & \textbf{34.57}\textsuperscript{\textcolor{red}{9.80}} & \textbf{39.02}\textsuperscript{\textcolor{red}{13.41}} & 37.20\textsuperscript{\textcolor{red}{14.03}}          & 28.05\textsuperscript{\textcolor{red}{3.66}}           \\
\codem-Java       & 35.37\textsuperscript{\textcolor{red}{8.54}}           & 33.54\textsuperscript{\textcolor{red}{9.15}}           & 32.30\textsuperscript{\textcolor{red}{3.73}}           & 29.63\textsuperscript{\textcolor{red}{4.94}}          & 31.10\textsuperscript{\textcolor{red}{5.49}}           & 37.80\textsuperscript{\textcolor{red}{14.63}}          & 27.44\textsuperscript{\textcolor{red}{3.05}}           \\
\codem-Go         & 35.98\textsuperscript{\textcolor{red}{9.15}}           & 33.54\textsuperscript{\textcolor{red}{9.15}}           & 31.68\textsuperscript{\textcolor{red}{3.11}}           & 30.25\textsuperscript{\textcolor{red}{5.56}}          & 34.15\textsuperscript{\textcolor{red}{8.54}}           & 35.98\textsuperscript{\textcolor{red}{12.81}}          & \textbf{32.32}\textsuperscript{\textcolor{red}{7.93}}  \\
\codem-HTML       & 31.71\textsuperscript{\textcolor{red}{4.88}}           & 33.54\textsuperscript{\textcolor{red}{9.15}}           & 32.30\textsuperscript{\textcolor{red}{3.73}}           & 25.93\textsuperscript{\textcolor{red}{1.24}}          & 28.66\textsuperscript{\textcolor{red}{3.05}}           & 38.41\textsuperscript{\textcolor{red}{15.24}}          & 28.05\textsuperscript{\textcolor{red}{3.66}}           \\
\codem-Mixed      & \textbf{43.29}\textsuperscript{\textcolor{red}{16.46}} & 37.20\textsuperscript{\textcolor{red}{12.81}}          & 37.89\textsuperscript{\textcolor{red}{9.32}}           & 32.10\textsuperscript{\textcolor{red}{7.41}}          & 37.80\textsuperscript{\textcolor{red}{12.19}}          & \textbf{39.63}\textsuperscript{\textcolor{red}{16.46}} & 29.27\textsuperscript{\textcolor{red}{4.88}}           \\
\bottomrule
\end{tabular}
\end{table}

\begin{table}
\centering
\caption{Pass$@1$ of StarCoder $15$B and \codem-Python. The numbers in \textcolor{red}{red} denote the absolute improvement compared to StarCoder 15B.}
\label{tab:results_15B}
\begin{tabular}{l|lllllll} 
\toprule
\multirow{2}{*}{\textbf{ Model}} & \multicolumn{7}{c}{\textbf{HumanEval-Multilingual}}                                                                                                                                                                                                                                                                                          \\
                                 & Python                                        & JavaScript                                    & TypeScript                                   & C                                            & C++                                           & Java                                          & Go                                             \\ 
\hline\hline
StarCoder 15B                    & 32.93                                         & 30.79                                         & 32.29                                        & 26.99                                        & 31.55                                         & 30.22                                         & 17.61                                          \\
\codem-Python     & 64.63\textsuperscript{\textcolor{red}{31.70}} & 47.56\textsuperscript{\textcolor{red}{16.77}} & 39.75\textsuperscript{\textcolor{red}{7.46}} & 35.19\textsuperscript{\textcolor{red}{9.20}} & 43.80\textsuperscript{\textcolor{red}{12.35}} & 48.17\textsuperscript{\textcolor{red}{17.95}} & 34.76\textsuperscript{\textcolor{red}{17.15}}  \\
\bottomrule
\end{tabular}
\end{table}

\section{Related Work}

Codex~\citep{codex} with 12-billion parameters is able to solve Python programming problems automatically.
This remarkable success triggered a significant buzz in both the academic and industrial realms.
Followed by Codex, a plenty of code LLMs are proposed, including AlphaCode~\citep{alphacode}, PaLM-Coder~\citep{palm}, CodeGen~\citep{codegen}, InCoder~\citep{incoder}, CodeGeeX~\citep{codegeex}, replit\footnote{\url{https://huggingface.co/replit/replit-code-v1-3b}}, CodeT5~\citep{codet5,codet5plus}, PyCodeGPT~\citep{cert}, SantaCoder~\citep{santacoder}, StarCoder~\citep{starcoder}, Code Llama~\citep{codellama}, and phi-1~\citep{phi1}.
These above models are trained on a large-scale code corpus and achieve impressive code generation performance.
During their pre-training, some models are trained on datasets of multilingual programming languages and then fine-tuned on a monolingual dataset to produce a more powerful specialist version. 
As for the instruction fine-tuning phase, WizardCoder~\citep{wizardcoder}, PanGu-Coder2~\citep{pangucoder2}, and Phind-CodeLlama\footnote{\url{https://huggingface.co/Phind/Phind-CodeLlama-34B-v1}} are proposed to bolster the capability of following instructions and further boost the code generation capability.
Yet, none of these aforementioned models explore the intricate interplay between different programming languages.
In this report, we therefore would like to investigate whether training code LLMs on monolingual data can bolster performance in other programming languages.

\section{Conclusion}

Our findings reveal that a monolingual training corpus can enhance the multilingual code generation capabilities of code LLMs via instruction tuning.
This highlights the intrinsic commonality and interconnectedness among multiple programming languages.
In our future work, we plan to delve into the reasons why multiple languages can enhance each other.
Also, we will explore how to leverage our findings to elevate code generation capabilities for these obscure or less-used programming languages by training on data from those popular ones.

\section*{Acknowledgements}
We would like to thank our colleagues for their valuable feedback and insights. Special thanks to An Fu (Huawei), Jingyang Zhao (Huawei), and Yuenan Guo (Huawei) for their constructive help throughout this research.

\bibliographystyle{unsrtnat}
\bibliography{references}

\end{document}